\documentclass[11pt,english]{article}
\usepackage[T1]{fontenc}
\usepackage[latin1]{inputenc}
\usepackage{babel}
\usepackage{setspace}
\makeatletter

\newcommand{\LyX}{L\kern-.1667em\lower.25em\hbox{Y}\kern-.125emX\spacefactor1000}

\makeatother

\begin{document}

\title{Distorted English Alphabet Identification:\\ An application of Difference Boosting
Algorithm.}

\author{Ninan Sajeeth Philip\thanks{
E-mail : nsp@stthom.ernet.in
} \protect\( \, \protect \)and K. Babu Joseph\thanks{
smanager@giasmd01.vsnl.net.in
}\\ Cochin University of Science and Technology, Kochi.}

\maketitle
\begin{abstract}
The difference boosting algorithm is used on the letters dataset from the UCI
repository to classify distorted raster images of English alphabets. In contrast
to rather complex networks, the difference boosting is found to produce comparable
or better classification efficiency on this complex problem. With a complete
set of 16000 training examples and two chances for making the correct prediction,
the network classified correctly in 98.35\% instances of the complete 20,000
examples. The accuracy in the first chance was 94.1\%.

\end{abstract}

\section{Introduction}

Learning is a process that involves identification and distinction of objects.
A child learns by examples. In the early days of learning, a child identifies
the proximity of his mother by smell or sound or body temperature. Thus it is
often not easy for him to distinguish between his mother, grandmother or mother's
sister in the first few months of his learning process. But as he grows and
his brain develops, he starts identifying the differences in each of his observations.
He learns to differentiate colours, between flowers and leaves, between friends
and enemies. This is reflected in his responses too. From the binary expressions
of smiling and weeping, different expressions appear in his reactions. He learns
to read the expressions on the faces of others and also develops the skill to
manage it. Difference boosting is an attempt to implement these concepts of
child learning into machine code. In the first round it picks up the unique
features in each observation using the naive Bayes' theorem, something that
is shown to be very similar to the functions of animal brain in object identification.
In the second step it attempts to boost the differences in these features that
enables one to differentiate almost similar objects. This is analogous to the
contention of the child that something that looks like a human but having a
tail is a monkey. The feature 'having a tail' gets boosted even if all the other
features are strikingly similar. Or, this is how the mother is able to effortlessly
distinguish her identical twin children or the artist is able to alter the expressions
on the face of his portrait with a few strokes of disjoint line segments. Classic
examples to these include the lighting effects on the facial expressions of
a sculpture. In all these examples, the brain does not appear to keep the details
but only those unique differences required to identify the objects. 

In complex problems, such as the distorted English alphabet detection example
discussed in this paper, the selection of a complete training set is not trivial.
We thus device a set of rules to identify the examples from a dataset that form
the complete training set for the network. The rules we follow in the said example
are:

1. If the network classify an object incorrectly, but with a high probability
of around 90\% or above, the example could be a new possibility and should be
included in the training set.

2. The network is allowed to produce a second guess on the possible class of
an object when it fails in the first prediction. If this is a correct guess
and if the difference between the degree of confidence between the first and
the second guess is greater than 90\% or less than 2\%, again the example is
assumed to be a new possibility or is in the vicinity of boostable examples
and is added to the training set.

The underlying logic in these rules are simple. Psychologists point out that
the human expert also develops his skill primarily based on past experiences
rather than on logical deduction or symbolic reasoning\cite{charness81}. We
also expect a similar situation in the learning process and assume that if an
example is incorrectly classified with a confidence higher than 90\%, there
could be two possible reasons for this. One possibility is that the network
is unaware of the existence of that example in the stated class. The other possibility
is that the features used are identical to that of an object from another class.
In this case, classification of the object into its actual class is difficult
without additional information. Assuming that the reason for the misclassification
is that such an sample is not known in the training set, we add that sample
also to the training set. In the second rule, we take the difference of the
confidence levels since we want to identify new examples and to tackle the border
problem that makes it difficult for the network to identify the exact class
based on the limited information content in the given features. The first condition
picks up the new examples while the second condition picks up the border examples.
These are the so-called 'difficult problems' in the learning process. 

One word of caution here is that the purpose of these rules are just to pickup
a complete set of examples in the training set, which is a pre-requisite of
any probability dependent classification problem. Once this dataset is generated,
the training process is done on this dataset and the testing is done on the
entire dataset and also on the independent test set. A good classification is
when the classification accuracy in both these cases are more or less the same.

\section{Naive Bayesian learning}

Each object has some characteristic features that enables the human brain to
identify and characterize them. These feature values or attributes might be
either associated to the object by a logical AND or a logical OR relation. The
total probability of a system with feature values associated by the logical
OR relation is the sum of the individual probabilities. Naive Bayesian classifiers
handles this situation by assigning probability distribution values for each
attribute separately. If on the otherhand, the attributes are associated by
a logical AND relation, meaning that each attribute value should be somewhere
around a stipulated value simultaneously, then the total probability is given
by the product of the individual probabilities of the attribute values. 

Now, the naive Bayesian classifier assumes that it is possible to assign some
degree of confidence to each attribute value of an example while attempting
to classify an object. Assume that the training set is complete with \( K \)
different known discrete classes. Then a statistical analysis should assign
a maximal value of the conditional probability \( P(C_{k}\mid U) \) for the
actual class \( C_{k} \) of the example. By Bayes' rule this probability may
be computed as :
\[
P(C_{k}\mid U)=\frac{P(U\mid C_{k})\, P(C_{k})}{\sum _{K}P(U\mid C_{k})\! P(C_{k})}\]

\( P(C_{k}) \) is also known as the background probability. \( P(U\mid C_{k}) \)
is given by the product of the probabilities due to individual attributes. That
is: 
\[
P(U\mid C_{k})=\prod _{m}P(U_{m}\mid C_{k})\]
 Following the axioms of set theory, one can compute \( P(U_{m}\mid C_{k}) \)
as \( P(U_{m}\cap C_{k}) \). This is nothing but the ratio of the total count
of the attribute value \( U_{m} \) in class \( C_{k} \) to the number of examples
in the entire training set. Thus naive Bayesian classifiers complete a training
cycle much faster than perceptrons or feed-forward neural networks.

\section{Difference Boosting\label{boosting} }

Boosting is an iterative process by which the network upweights misclassified
examples in a training set until it is correctly classified. The Adaptive Boosting
(AdaBoost) algorithm of Freund and Schapire \cite{Freund95, Freund97} attempts
the same thing. In this paper, we present a rather simple algorithm for boosting.
The structure of our network is identical to AdaBoost in that it also modifies
a weight function. Instead of computing the error in the classification as the
total error produced in the training set, we take each misclassified example
and apply a correction to its weight based on its own error. Also, instead of
upweighting an example, our network upweights the weight associated to the probability
\( P(U_{m}\mid C_{k}) \) of each attribute of the example. Thus the modified
weight will affect all the examples that have the same attribute value even
if its other attributes are different. During the training cycle, there is a
competitive update of attribute weights to reduce the error produced by each
example. It is expected that at the end of the training epoch the weights associated
to the probability function of each attribute will stabilize to some value that
produces the minimum error in the entire training set. Identical feature values
compete with each other and the differences get boosted up. Thus the classification
becomes more and more dependent on the differences rather than on similarities.
This is analogous to the way in which the human brain differentiates between
almost similar objects by sight, like for example, rotten tomatoes from a pile
of good ones. 

Let us consider a misclassified example in which \( P_{k} \) represent the
computed probability for the actual class \( k \) and \( P_{k}^{*} \) that
for the wrongly represented class. Our aim is to push the computed probability
\( P_{k} \) to some value greater than \( P^{*}_{k} \). In our network, this
is done by modifying the weight associated to each \( P(U_{m}\mid C_{k}) \)
of the misclassified item by the negative gradient of the error, i.e. \( \Delta W_{m}=\alpha \left[ 1-\frac{P_{k}}{P^{*}_{k}}\right]  \).
Here \( \alpha  \) is a constant which determines the rate at which the weight
changes. The process is repeated until all items are classified correctly or
a predefined number of rounds completes.

\section{The classifier network.}

Assuming that the occurrences of the classes are equally probable, we start
with a flat prior distribution of the classes ,i.e. \( P(C_{k})=\frac{1}{N} \).
This might appear unrealistic, since this is almost certain to be unequal in
most practical cases. The justification is that since \( P(C_{K}) \) is also
a weighting function, we expect this difference also to be taken care of by
the connection weights during the boosting process. The advantage on the otherhand
is that it avoids any assumptions on the training set regarding the prior estimation.
Now, the network presented in this paper may be divided into three units. The
first unit computes the Bayes' probability for each of the training examples.
If there are \( M \) number of attributes with values ranging from \( m_{min} \)
to \( m_{max} \) and belonging to one of the \( K \) discrete classes, we
first construct a grid of equal sized bins for each \( k \) with columns representing
the attributes and rows their values. Thus a training example \( S_{i} \) belonging
to a class \( k \) and having one of its attributes \( l \) with a value \( m \)
will fall into the bin \( B_{klm} \) for which the Euclidean distance between
the center of the bin and the attribute value is a minimum. The number of bins
in each row should cover the range of the attributes from \( m_{min} \) to
\( m_{max} \). It is observed that there exist an optimum number of bins that
produce the maximum classification efficiency for a given problem. For the time
being, it is computed by trial and error. Once this is set, the training process
is simply to distribute the examples in the training sets into their respective
bins. After this, the number of attributes in each bin \( i \) for each class
\( k \) is counted and this gives the probability \( P(U_{m}\mid C_{k}) \)
of the attribute \( m \) with value \( U_{m}\equiv i \) for the given \( C_{k}=k \).
The basic difference of this new formalism with that of the popular gradient
descent backpropagation algorithm and similar Neural Networks is that, here
the distance function is the distance between the probabilities, rather than
the feature magnitudes. Thus the new formalism can isolate overlapping regions
of the feature space more efficiently than standard algorithms. 

The naive Bayesian learning fails when the data set represent an XOR like feature.
To overcome this, associated to each row of bins of the attribute values we
put a tag that holds the minimum and maximum values of the other attributes
in the data example. This tag acts as a level threshold window function. In
our example, if an attribute value in the example happens to be outside the
range specified in the tag, then the computed \( P(U_{m}\mid C_{k}) \) of that
attribute is reduced to one-forth of its actual value (gain of 0.25). Applying
such a simple window enabled the network to handle the XOR kind of problems
efficiently.

The second unit in the network is the gradient descent boosting algorithm. To
do this, each of the probability components \( P(U_{m}\mid C_{k}) \) is amplified
by a connection weight before computing \( P(U\mid C_{k}) \). Initially all
the weights are set to unity. For a correctly classified example, \( P(U\mid C_{k}) \)
will be a maximum for the class specified in the training set. For the misclassified
items, we increment its weight by a fraction \( \Delta W_{m} \). The training
set is read repeatedly for a few rounds and in each round the connection weights
of the misclassified items are incremented by \( \Delta W_{m}=\alpha \left[ 1-\frac{P_{k}}{P^{*}_{k}}\right]  \)
as explained in section \ref{boosting}, until the item is classified correctly. 

The third unit computes \( P(C_{k}\mid U) \) as :

\[
P(C_{k}\mid U)=\frac{\prod _{m}P(U_{m}\mid C_{k})\! W_{m}}{\sum _{K}\prod _{m}P(U_{m}\mid C_{k})\! W_{m}}\]
 If this is a maximum for the class given in the training set, the network is
said to have learned correctly. The wrongly classified items are re-submitted
to the boosting algorithm in the second unit.

\section{Results on the letters dataset}

The letters dataset consists of 20,000 unique letter images generated randomly
distorting pixel images of the 26 uppercase letters from 20 different commercial
fonts. Details of these dataset may be found in \cite{slate91}. The parent font
represented a full range of character types including script, italic, serif
and Gothic. The features of each of the 20,000 characters were summarized in
terms of 16 primitive numerical attributes. The attributes are\cite{slate91}:

\begin{enumerate}

\item The horizontal position, counting pixels from the left edge of the image, of the center of the smallest rectangular box that can be drawn with all "on" pixels inside the box. 

\item The vertical position, counting pixels from the bottom, of the box. 

\item The width, in pixels, of the box. 

\item The height, in pixels, of the box. 

\item The total number of "on" pixels in the charecter image. 

\item The mean horizontal position of all "on" pixels relative to the center of the box and divided by the width of the box. This feature has a negative value if the image is "left heavy" as would be the case for the letter L. 

\item The mean vertical position of all "on" pixels relative to the center of the box and divided by the height of the box. 

\item The mean squared value of the horizontal pixel distances as measured in 6 above. This attribute will have a higher value for images whose pixels are more widely separated in the horizontal direction as would be the case for the letters W and M. 

\item The mean squared value of the vertical pixel distances as measured in 7 above. 

\item The mean product of the horizontal and vertical distances for each "on" pixel as measured in 6 and 7 above. This attribute has a positive value for diagonal lines that run from bottom left to top right and a negative value for diagonal lines from top left to bottom right. 

\item The mean value of the squared horizontal distance times the vertical distance for each "on" pixel. This measures the correlation of the horizontal variance with the vertical position. 

\item The mean value of the squared vertical distance times the horizontal distance for each "on" pixel. This measures the correlation of the vertical variance with the horizontal position. 

\item The mean number of edges (an "on" pixel immediately to the right of either an "off" pixel or the image boundary) encountered when making systematic scans from left to right at all vertical positions within the box. This measure distinguishes between letters like "W" or "M" and letters like "I" or "L". 

\item The sum of the vertical positions of edges encountered as measured in 13 above. This feature will give a higher value if there are more edges at the top of the box, as in the letter "Y". 

\item The mean number of edges ( an "on" pixel immediately above either an "off" pixel or the image boundary ) encountered when making systematic scans of the image from bottom to top over all horizontal positions within the box. \item The sum of horizontal positions of edges encountered as measured in 15 above.

\end{enumerate}

Using a Holland-style adaptive classifier and a training set of 16,000 examples,
the classifier accuracy reported on this dataset\cite{slate91} is a little over
80\%. The naive Bayesian classifier\cite{zheng99} produces an error rate of
25.26\% while when boosted with AdaBoost reduces the error to 24.12\%. Using
AdaBoost on the C4.5 algorithm\cite{schapire97} could reduce the error to 3.1\%
on the testset. However the computational power required over 100 machines to
generate the tree structure\cite{bengio} for its effectuation. A fully connected
MLP with 16-70-50-26 topology\cite{bengio} gave an error of 2.0\% with AdaBoost
and required 20 machines to implement the system. 

The proposed algorithm on a single Celeron processor running at 300MHz and Linux
6.0 attained an error rate of 14.2 \% on the independent testset of 4000 examples
in less than 15 minutes. Applying the rules mentioned in section one resulted
in an overall error of 5.9 \% on 20,000 examples. With two chances, the error
went as low as 1.65 \%. The result is promising taking into consideration the
low computational power required by the system. Since bare character recognition
of this rate can be improved with other techniques such as grammer and cross-word
lookup methods, we expect a near 100 \% recognition rate on such systems. Further
work is to look into this aspect in detail.

\section{Conclusion}

Bayes' rule on how the degree of belief should change on the basis of evidences
is one of the most popular formalism for brain modeling. In most implementations,
the degree of belief is computed in terms of the degree of agreement to some
known criteria. However, this has the disadvantage that some of the minor differences
might be left unnoticed by the classifier. We thus device a classifier that
pays more attention to differences rather than similarities in identifying the
classes from a dataset. In the training epoch, the network identifies the apparent
differences and magnify them to separate out classes. We applied the classifier
on many practical problems and found that this makes sense. The application
of the method on the letter dataset produced an error as low as 1.65 \% when
two chances were given to make the prediction. Further study is to look into
the application of other language recognition techniques in conjunction with
the network. 

{\noindent}{\bf{Acknowledgment}}

Ninan Sajeeth Philip would like to thank David J Slate (Pattern Recognition
Group, Odesta Corporation, Evanston) for sending him the original of their paper
and also for the useful comments and discussions.


\begin{thebibliography}{}
\bibitem{charness81}Charness, N., Aging and skilled problem-solving, Journal of experimental psychology:General,
110, 21-38, 1981.
\bibitem{Freund95}Yoav Freund and Robert Schapire, E., A decision-theoretic generalization of
on-line learning and an application to boosting. in \textit{Proceedings of Second
European Conference on Computational Learning Theory}, 23-37,1995
\bibitem{Freund97}Yoav Freund and Robert Schapire, E., A decision-theoretic generalization of
on-line learning and an application to boosting. \textit{Journal of Computer
and System Sciences}, 55(1), 119-139, 1997.
\bibitem{slate91}Peter W Frey and David J Slate, Letter Recognition using Holland-style adaptive
classifiers. \textit{Machine Learning,} 6,161-182 ,1991.
\bibitem{zheng99}Kai Ming Ting and Zijian Zheng, Proceedings of PAKDD'99, Springer-Verlag, 1999.
\bibitem{schapire97}Robert E. Schapire, Yoav Freund, Peter Bartlett and Wee Sun Lee. Boosting the
margin: A new explanation for the effectiveness of voting methods, Machine Learning:
Proceedings of the fourteenth International Conference,1997.
\bibitem{bengio}Holger Schwenk and Yoshua Bengio, Adaptive Boosting of Neural Networks for Character
Recognition, Technical report \#1072, Department d' Informatique et Recherche
Ope'rationnelle, Universite' de Montre'al, Montreal, Qc H3C-3J7, Canada.
\end{thebibliography}
\end{document}